\PassOptionsToPackage{unicode}{hyperref}
\PassOptionsToPackage{hyphens}{url}
\PassOptionsToPackage{dvipsnames,svgnames,x11names}{xcolor}
\documentclass[
]{article}

\usepackage{amsmath,amssymb}
\usepackage{lmodern}
\usepackage{iftex}
\ifPDFTeX
  \usepackage[T1]{fontenc}
  \usepackage[utf8]{inputenc}
  \usepackage{textcomp} 
\else 
  \usepackage{unicode-math}
  \defaultfontfeatures{Scale=MatchLowercase}
  \defaultfontfeatures[\rmfamily]{Ligatures=TeX,Scale=1}
\fi
\IfFileExists{upquote.sty}{\usepackage{upquote}}{}
\IfFileExists{microtype.sty}{
  \usepackage[]{microtype}
  \UseMicrotypeSet[protrusion]{basicmath} 
}{}
\makeatletter
\@ifundefined{KOMAClassName}{
  \IfFileExists{parskip.sty}{%
    \usepackage{parskip}
  }{
    \setlength{\parindent}{0pt}
    \setlength{\parskip}{6pt plus 2pt minus 1pt}}
}{
  \KOMAoptions{parskip=half}}
\makeatother
\usepackage{xcolor}
\ifx\paragraph\undefined\else
  \let\oldparagraph\paragraph
  \renewcommand{\paragraph}[1]{\oldparagraph{#1}\mbox{}}
\fi
\ifx\subparagraph\undefined\else
  \let\oldsubparagraph\subparagraph
  \renewcommand{\subparagraph}[1]{\oldsubparagraph{#1}\mbox{}}
\fi

\usepackage{longtable,booktabs,array}
\usepackage{calc} 
\usepackage{etoolbox}
\makeatletter
\patchcmd\longtable{\par}{\if@noskipsec\mbox{}\fi\par}{}{}
\makeatother
\IfFileExists{footnotehyper.sty}{\usepackage{footnotehyper}}{\usepackage{footnote}}
\makesavenoteenv{longtable}
\usepackage{graphicx}
\makeatletter
\def\maxwidth{\ifdim\Gin@nat@width>\linewidth\linewidth\else\Gin@nat@width\fi}
\def\maxheight{\ifdim\Gin@nat@height>\textheight\textheight\else\Gin@nat@height\fi}
\makeatother
\setkeys{Gin}{width=\maxwidth,height=\maxheight,keepaspectratio}
\makeatletter
\def\fps@figure{htbp}
\makeatother
\newlength{\cslhangindent}
\newlength{\csllabelwidth}
\newlength{\cslentryspacingunit} 
\newenvironment{CSLReferences}[2] 
 {
  \setlength{\parindent}{0pt}
  \ifodd #1
  \let\oldpar\par
  \def\par{\hangindent=\cslhangindent\oldpar}
  \fi
  \setlength{\parskip}{#2\cslentryspacingunit}
 }%
 {}
\usepackage{calc}

\usepackage{arxiv}
\usepackage{orcidlink}
\usepackage{amsmath}
\usepackage[T1]{fontenc}
\makeatletter
\@ifpackageloaded{caption}{}{\usepackage{caption}}
\AtBeginDocument{%
\ifdefined\contentsname
  \renewcommand*\contentsname{Table of contents}
\else
  \newcommand\contentsname{Table of contents}
\fi
\ifdefined\listfigurename
  \renewcommand*\listfigurename{List of Figures}
\else
  \newcommand\listfigurename{List of Figures}
\fi
\ifdefined\listtablename
  \renewcommand*\listtablename{List of Tables}
\else
  \newcommand\listtablename{List of Tables}
\fi
\ifdefined\figurename
  \renewcommand*\figurename{Figure}
\else
  \newcommand\figurename{Figure}
\fi
\ifdefined\tablename
  \renewcommand*\tablename{Table}
\else
  \newcommand\tablename{Table}
\fi
}
\@ifpackageloaded{float}{}{\usepackage{float}}
\floatstyle{ruled}
\@ifundefined{c@chapter}{\newfloat{codelisting}{h}{lop}}{\newfloat{codelisting}{h}{lop}[chapter]}
\floatname{codelisting}{Listing}

\makeatother
\makeatletter
\@ifpackageloaded{caption}{}{\usepackage{caption}}
\@ifpackageloaded{subcaption}{}{\usepackage{subcaption}}
\makeatother
\makeatletter
\@ifpackageloaded{tcolorbox}{}{\usepackage[many]{tcolorbox}}
\makeatother
\makeatletter
\@ifundefined{shadecolor}{\definecolor{shadecolor}{rgb}{.97, .97, .97}}
\makeatother
\ifLuaTeX
  \usepackage{selnolig}  
\fi
\IfFileExists{bookmark.sty}{\usepackage{bookmark}}{\usepackage{hyperref}}
\IfFileExists{xurl.sty}{\usepackage{xurl}}{} 
\hypersetup{
  pdftitle={TextDescriptives: A Python package for calculating a large variety of metrics from text},
  pdfauthor={Lasse Hansen; Ludvig Renbo Olsen; Kenneth Enevoldsen},
  pdfkeywords={Python, natural language processing, spacy, feature
extraction},
  colorlinks=true,
  linkcolor={blue},
  filecolor={Maroon},
  citecolor={Blue},
  urlcolor={Blue},
  pdfcreator={LaTeX via pandoc}}

\newcommand{\runninghead}{A Preprint }
\renewcommand{\runninghead}{TextDescriptives }
\title{TextDescriptives: A Python package for calculating a large
variety of metrics from text}
\author{
\textbf{Lasse Hansen}~\orcidlink{0000-0003-1113-4779}\\Department of
Affective Disorders - Psychiatry\\Aarhus University Hospital\\Aarhus,
Denmark\\\\\\\\
\textbf{Ludvig Renbo Olsen}~\orcidlink{0009-0006-6798-7454}\\Department
of Molecular Medicine\\Aarhus University\\Aarhus, Denmark\\\\\\\\
\textbf{Kenneth Enevoldsen}~\orcidlink{0000-0001-8733-0966}\\Center for
Humanities Computing\\Aarhus University\\Aarhus, Denmark\\}
\date{}
\begin{document}
\maketitle
\begin{abstract}
TextDescriptives is a Python package for calculating a large variety of
statistics from text. It is built on top of spaCy and can be easily
integrated into existing workflows. The package has already been used
for analysing the linguistic stability of clinical texts, creating
features for predicting neuropsychiatric conditions, and analysing
linguistic goals of primary school students. This paper describes the
package and its features.
\end{abstract}
{\bfseries \emph Keywords}
\def\sep{\textbullet\ }
Python \sep natural language processing \sep spacy \sep 
feature extraction

\ifdefined\Shaded\renewenvironment{Shaded}{\begin{tcolorbox}[breakable, boxrule=0pt, sharp corners, interior hidden, borderline west={3pt}{0pt}{shadecolor}, frame hidden, enhanced]}{\end{tcolorbox}}\fi

\hypertarget{summary}{%
\section{Summary}\label{summary}}

Natural language processing (NLP) tasks often require a thorough
understanding and description of the corpus. Document-level metrics can
be used to identify low-quality data, assess outliers, or understand
differences between groups. Further, text metrics have long been used in
fields such as the digital humanities where e.g.~metrics of text
complexity are commonly used to analyse, understand and compare text
corpora. However, extracting complex metrics can be an error-prone
process and is rarely rigorously tested in research implementations.
This can lead to subtle differences between implementations and reduces
the reproducibility of scientific results.

\texttt{TextDescriptives} offers a simple and modular approach to
extracting both simple and complex metrics from text. It achieves this
by building on the \texttt{spaCy} framework (Honnibal et al. 2020). This
means that \texttt{TextDescriptives} can easily be integrated into
existing workflows while leveraging the efficiency and robustness of the
\texttt{spaCy} library. The package has already been used for analysing
the linguistic stability of clinical texts (Hansen et al. 2022),
creating features for predicting neuropsychiatric conditions (Hansen et
al. 2023), and analysing linguistic goals of primary school students
(Tannert 2023).

\hypertarget{statement-of-need}{%
\section{Statement of need}\label{statement-of-need}}

Computational text analysis is a broad term that refers to the process
of analyzing and understanding text data. This often involves
calculating a set of metrics that describe relevant properties of the
data. Dependent on the task at hand, this can range from simple
descriptive statistics related to e.g.~word or sentence length to
complex measures of text complexity, coherence, or quality. This often
requires drawing on multiple libraries and frameworks or writing custom
code. This can be time-consuming and prone to bugs, especially with more
complex metrics.

\texttt{TextDescriptives} seeks to unify the extraction of
document-level metrics, in a modular fashion. The integration with spaCy
allows the user to seamlessly integrate \texttt{TextDescriptives} in
existing pipelines as well as giving the \texttt{TextDescriptives}
package access to model-based metrics such as dependency graphs and
part-of-speech tags. The ease of use and the variety of available
metrics allows researchers and practitioners to extend the granularity
of their analyses within a tested and validated framework.

Implementations of the majority of the metrics included in
\texttt{TextDescriptives} exist, but none as feature complete. The
\texttt{textstat} library (Ward 2022) implements the same readability
metrics, however, each metric has to be extracted one at a time with no
interface for multiple extractions. \texttt{spacy-readability}
(Holtzscher 2019) adds readability metrics to \texttt{spaCy} pipelines,
but does not work for new versions of \texttt{spaCy}
(\textgreater=3.0.0). The \texttt{textacy} (DeWilde 2021) package has
some overlap with \texttt{TextDescriptives}, but with a different focus.
\texttt{TextDescriptives} focuses on document-level metrics, and
includes a large number of metrics not included in \texttt{textacy}
(dependency distance, coherence, and quality), whereas \texttt{textacy}
includes components for preprocessing, information extraction, and
visualization that are outside the scope of \texttt{TextDescriptives}.
What sets \texttt{TextDescriptives} apart is the easy access to
document-level metrics through a simple user-facing API and exhaustive
documentation.

\hypertarget{features-functionality}{%
\section{Features \& Functionality}\label{features-functionality}}

\texttt{TextDescriptives} is a Python package and provides the following
\texttt{spaCy} pipeline components:
\texttt{textdescriptives.descriptive\_stats}: Calculates the total
number of tokens, number of unique tokens, number of characters, and the
proportion of unique tokens, as well as the mean, median, and standard
deviation of token length, sentence length, and the number of syllables
per token. \texttt{textdescriptives.readability}: Calculates the
Gunning-Fog index, the SMOG index, Flesch reading ease, Flesch-Kincaid
grade, the Automated Readability Index, the Coleman-Liau index, the Lix
score, and the Rix score.
\texttt{textdescriptives.dependency\_distance}: Calculates the mean and
standard deviation of the dependency distance (the average distance
between a word and its head word), and the mean and the standard
deviation of the proportion adjacent dependency relations on the
sentence level. \texttt{textdescriptives.pos\_proportions}: Calculates
the proportions of all part-of-speech tags in the documents.
\texttt{textdescriptives.coherence}: Calculates the first- and
second-order coherence of the document based on word embedding
similarity between sentences. \texttt{textdescriptives.quality}:
Calculates the text-quality metrics proposed in Rae et al. (2022) and
Raffel et al. (2020). These measures can be used for filtering out
low-quality text prior to model training or text analysis. These include
heuristics such as the number of stop words, ratio of words containing
alphabetic characters, proportion of lines ending with an ellipsis,
proportion of lines starting with a bullet point, ratio of symbols to
words, and whether the document contains a specified string
(e.g.~``lorem ipsum''), as well as repetitious text metrics such as the
proportion of lines that are duplicates, the proportion of paragraphs in
a document that are duplicates, the proportion of n-gram duplicates, and
the proportion of characters in a document that are contained within the
top n-grams.

All the components can be added to an existing \texttt{spaCy} pipeline
with a single line of code, and jointly extracted to a dataframe or
dictionary with a single call to
\texttt{textdescriptives.extract\_\{df\textbar{}dict\}(doc)}.

\hypertarget{example-use-cases}{%
\section{Example Use Cases}\label{example-use-cases}}

Descriptive statistics can be used to summarize and understand data,
such as by exploring patterns and relationships within the data, getting
a better understanding of the data set, or identifying any changes in
the distribution of the data. Readability metrics, which assess the
clarity and ease of understanding of written text, have a variety of
applications, including the design of educational materials and the
improvement of legal or technical documents (DuBay 2004). Dependency
distance can be used as a measure of language comprehension difficulty
or of sentence complexity and has been used for analysing properties of
natural language or for similar purposes as readability metrics (Gibson
et al. 2019; Liu 2008). The proportions of different parts of speech in
a document have been found to be predictive of certain mental disorders
and can also be used to assess the quality and complexity of text (Tang
et al. 2021). Semantic coherence, or the logical connection between
sentences, has primarily been used in the field of computational
psychiatry to predict the onset of psychosis or schizophrenia (Parola et
al. 2022; Bedi et al. 2015), but it also has other applications in the
digital humanities. Measures of text quality are useful cleaning and
identifying low-quality data (Rae et al. 2022; Raffel et al. 2020).

\hypertarget{target-audience}{%
\section{Target Audience}\label{target-audience}}

The package is mainly targeted at NLP researchers and practitioners. In
particular, researchers from fields new to NLP such as the digital
humanities and social sciences as researchers might benefit from the
readability metrics as well as the more complex, but highly useful,
metrics such as coherence and dependency distance.

\hypertarget{acknowledgements}{%
\section{Acknowledgements}\label{acknowledgements}}

The authors thank the
\href{https://github.com/HLasse/TextDescriptives/graphs/contributors}{contributors}
of the package including Martin Bernstorff for his work on the
part-of-speech component, and Frida Hæstrup and Roberta Rocca for
important fixes. The authors would also like to Dan Sattrup Nielsen for
helpful reviews on early iterations of the text quality implementations.

\hypertarget{references}{%
\section*{References}\label{references}}
\addcontentsline{toc}{section}{References}

\hypertarget{refs}{}
\begin{CSLReferences}{1}{0}
\leavevmode\vadjust pre{\hypertarget{ref-bedi_automated_2015}{}}%
Bedi, Gillinder, Facundo Carrillo, Guillermo A. Cecchi, Diego Fernández
Slezak, Mariano Sigman, Natália B. Mota, Sidarta Ribeiro, Daniel C.
Javitt, Mauro Copelli, and Cheryl M. Corcoran. 2015. {``Automated
Analysis of Free Speech Predicts Psychosis Onset in High-Risk Youths.''}
\emph{Npj Schizophrenia} 1 (1): 1--7.
\url{https://doi.org/10.1038/npjschz.2015.30}.

\leavevmode\vadjust pre{\hypertarget{ref-dewilde_textacy_2021}{}}%
DeWilde, Burton. 2021. \emph{Textacy: {NLP}, Before and After {spaCy}}
(version 0.12.0). \url{https://github.com/chartbeat-labs/textacy}.

\leavevmode\vadjust pre{\hypertarget{ref-dubay_principles_2004}{}}%
DuBay, William H. 2004. {``The Principles of Readability.''}
\emph{Online Submission}.

\leavevmode\vadjust pre{\hypertarget{ref-gibson_how_2019}{}}%
Gibson, Edward, Richard Futrell, Steven P. Piantadosi, Isabelle
Dautriche, Kyle Mahowald, Leon Bergen, and Roger Levy. 2019. {``How
Efficiency Shapes Human Language.''} \emph{Trends in Cognitive Sciences}
23 (5): 389--407.

\leavevmode\vadjust pre{\hypertarget{ref-hansen_lexical_2022}{}}%
Hansen, Lasse, Kenneth Enevoldsen, Martin Bernstorff, Erik Perfalk,
Andreas A. Danielsen, Kristoffer L. Nielbo, and Søren D. Østergaard.
2022. {``Lexical Stability of Psychiatric Clinical Notes from Electronic
Health Records over a Decade.''} \emph{{medRxiv}}, January,
2022.09.05.22279610. \url{https://doi.org/10.1101/2022.09.05.22279610}.

\leavevmode\vadjust pre{\hypertarget{ref-hansen_automated_2023}{}}%
Hansen, Lasse, Roberta Rocca, Arndis Simonsen, Alberto Parola, Vibeke
Bliksted, Nicolai Ladegaard, Dan Bang, et al. 2023. {``Automated Speech-
and Text-Based Classification of Neuropsychiatric Conditions in a
Multidiagnostic Setting,''} no. {arXiv}:2301.06916 (January).
\url{https://doi.org/10.48550/arXiv.2301.06916}.

\leavevmode\vadjust pre{\hypertarget{ref-holtzscher_spacy-readability_2019}{}}%
Holtzscher, Michael. 2019. \emph{Spacy-Readability: {spaCy} Pipeline
Component for Adding Text Readability Meta Data to Doc Objects.}
(version 1.4.1).

\leavevmode\vadjust pre{\hypertarget{ref-honnibal_spacy_2020}{}}%
Honnibal, Matthew, Ines Montani, Sofie Van Landeghem, and Adriane Boyd.
2020. \emph{{spaCy}: Industrial-Strength Natural Language Processing in
Python}. \url{https://doi.org/10.5281/zenodo.1212303}.

\leavevmode\vadjust pre{\hypertarget{ref-liu_dependency_2008}{}}%
Liu, Haitao. 2008. {``Dependency Distance as a Metric of Language
Comprehension Difficulty.''} \emph{Journal of Cognitive Science} 9 (2):
159--91. \url{https://doi.org/10.17791/jcs.2008.9.2.159}.

\leavevmode\vadjust pre{\hypertarget{ref-parola_speech_2022}{}}%
Parola, Alberto, Jessica Mary Lin, Arndis Simonsen, Vibeke Bliksted,
Yuan Zhou, Huiling Wang, Lana Inoue, Katja Koelkebeck, and Riccardo
Fusaroli. 2022. {``Speech Disturbances in Schizophrenia: Assessing
Cross-Linguistic Generalizability of {NLP} Automated Measures of
Coherence.''} \emph{Schizophrenia Research}, August.
\url{https://doi.org/10.1016/j.schres.2022.07.002}.

\leavevmode\vadjust pre{\hypertarget{ref-rae_scaling_2022}{}}%
Rae, Jack W., Sebastian Borgeaud, Trevor Cai, Katie Millican, Jordan
Hoffmann, Francis Song, John Aslanides, et al. 2022. {``Scaling Language
Models: Methods, Analysis \& Insights from Training Gopher.''}
{arXiv}:2112.11446. {arXiv}.
\url{https://doi.org/10.48550/arXiv.2112.11446}.

\leavevmode\vadjust pre{\hypertarget{ref-raffel_exploring_2020}{}}%
Raffel, Colin, Noam Shazeer, Adam Roberts, Katherine Lee, Sharan Narang,
Michael Matena, Yanqi Zhou, Wei Li, and Peter J. Liu. 2020. {``Exploring
the Limits of Transfer Learning with a Unified Text-to-Text
Transformer.''} \emph{{arXiv}:1910.10683 {[}Cs, Stat{]}}, July.
\url{http://arxiv.org/abs/1910.10683}.

\leavevmode\vadjust pre{\hypertarget{ref-tang_natural_2021}{}}%
Tang, Sunny X., Reno Kriz, Sunghye Cho, Suh Jung Park, Jenna Harowitz,
Raquel E. Gur, Mahendra T. Bhati, Daniel H. Wolf, João Sedoc, and Mark
Y. Liberman. 2021. {``Natural Language Processing Methods Are Sensitive
to Sub-Clinical Linguistic Differences in Schizophrenia Spectrum
Disorders.''} \emph{Npj Schizophrenia} 7 (1): 1--8.
\url{https://doi.org/10.1038/s41537-021-00154-3}.

\leavevmode\vadjust pre{\hypertarget{ref-tannert_skriftsproglig_2023}{}}%
Tannert, Morten. 2023. {``Skriftsproglig Udvikling i Grundskolens
Danskfag.''} PhD thesis, Aarhus University.

\leavevmode\vadjust pre{\hypertarget{ref-ward_textstat_2022}{}}%
Ward, Alex. 2022. \emph{Textstat}. Textstat.
\url{https://github.com/textstat/textstat}.

\end{CSLReferences}

\end{document}